\documentclass{article}

\usepackage{microtype}
\usepackage{graphicx}
\usepackage{subfigure}
\usepackage{booktabs}

\usepackage{hyperref}

\usepackage{xcolor}
\usepackage{amsmath}
\usepackage{amsthm}
\usepackage{amsbsy,amsfonts,amssymb}
\usepackage{graphicx}
\usepackage{multirow}
\usepackage{wrapfig}
\usepackage{subfigure}
\usepackage{tikz}
\usepackage{comment}
\usepackage[shortlabels]{enumitem}

\usepackage{algorithm}
\usepackage{algorithmic}

\usepackage[page]{appendix}

\graphicspath{ {./figs/} }

\renewcommand{\algorithmicrequire}{\textbf{Input:}}
\renewcommand{\algorithmicensure}{\textbf{Output:}}

\def\0{\textbf{0}}
\def\1{\textbf{1}}

\def\c{\boldsymbol{c}}

\def\x{\boldsymbol{x}}
\def\y{\boldsymbol{y}}
\def\z{\boldsymbol{z}}

\def\xm{\boldsymbol{\xi}}
\def\ym{\boldsymbol{\eta}}
\def\am{\boldsymbol{\alpha}}

\def\A{\boldsymbol{A}}
\def\X{\boldsymbol{X}}

\def\cA{\mathcal{A}}

\def\cD{\mathcal{D}}

\renewcommand{\mathbf}{\boldsymbol}

\newcommand{\bb}{\mathbb}

\renewcommand{\Re}{{\mathbb R}}

\newcommand{\Z}{\bb Z}

\newtheorem{definition}{Definition}
\newtheorem{theorem}{Theorem}
\newtheorem{example}{Example}

\definecolor{mediumred-violet}{rgb}{0.73, 0.2, 0.52}

\definecolor{demphcolor}{RGB}{160,160,160}
\newcommand{\demph}[1]{\textcolor{demphcolor}{#1}}
\newcommand{\app}{\raise.17ex\hbox{$\scriptstyle\sim$}}

\definecolor{urlcolor}{RGB}{0,0,238}

\newenvironment{mytabular}[1][1]{%
  \renewcommand*{\arraystretch}{#1}%
  \tabular%
}{%
  \endtabular
}
\usepackage[accepted]{icml2020}
\icmltitlerunning{Deep Isometric Learning for Visual Recognition}

\begin{document}
\twocolumn[
\icmltitle{Deep Isometric Learning for Visual Recognition}

\begin{icmlauthorlist}
\icmlauthor{Haozhi Qi}{ucb}
\icmlauthor{Chong You}{ucb}
\icmlauthor{Xiaolong Wang}{ucb,ucsd}
\icmlauthor{Yi Ma}{ucb}
\icmlauthor{Jitendra Malik}{ucb}\\
\icmlauthor{\rm{\small{\url{https://haozhiqi.github.io/ISONet}}}}{}
\end{icmlauthorlist}

\icmlaffiliation{ucb}{UC Berkeley}
\icmlaffiliation{ucsd}{UC San Diego}
\icmlcorrespondingauthor{Haozhi Qi}{hqi@berkeley.edu}
\icmlkeywords{Machine Learning, Image Classification, Computer Vision}
\vskip 0.3in
]
\printAffiliationsAndNotice{}

\begin{abstract}
Initialization, normalization, and skip connections are believed to be three indispensable techniques for training very deep convolutional neural networks and obtaining state-of-the-art performance. This paper shows that deep vanilla ConvNets without normalization nor skip connections can also be trained to achieve surprisingly good performance on standard image recognition benchmarks. This is achieved by enforcing the convolution kernels to be near \emph{isometric} during initialization and training, as well as by using a variant of ReLU that is shifted towards being \emph{isometric}. Further experiments show that if combined with skip connections, such near isometric networks can achieve performances on par with (for ImageNet) and better than (for COCO) the standard ResNet, even without normalization at all. Our code is available at \url{https://github.com/HaozhiQi/ISONet}.
\end{abstract}

\section{Introduction}

Convolutional Neural Networks (ConvNets) have achieved phenomenal success in computer vision~\cite{krizhevsky2012imagenet,simonyan2014very,szegedy2015going,he2015delving,ioffe2015batch,he2016deep,xie2017aggregated,huang2017densely}.
While shallow ConvNets with a few layers have existed for decades~\cite{lecun1989backpropagation, denker1989neural, lecun1990handwritten, lecun1998gradient}, it is only until recently that networks with hundreds or even thousands of layers can be effectively trained.
Such \emph{deep} ConvNets are able to learn sophisticated decision rules for complex practical data, therefore are usually indispensable for obtaining state-of-the-art performance.

\begin{figure}[t]
\centering
\includegraphics[width=\linewidth]{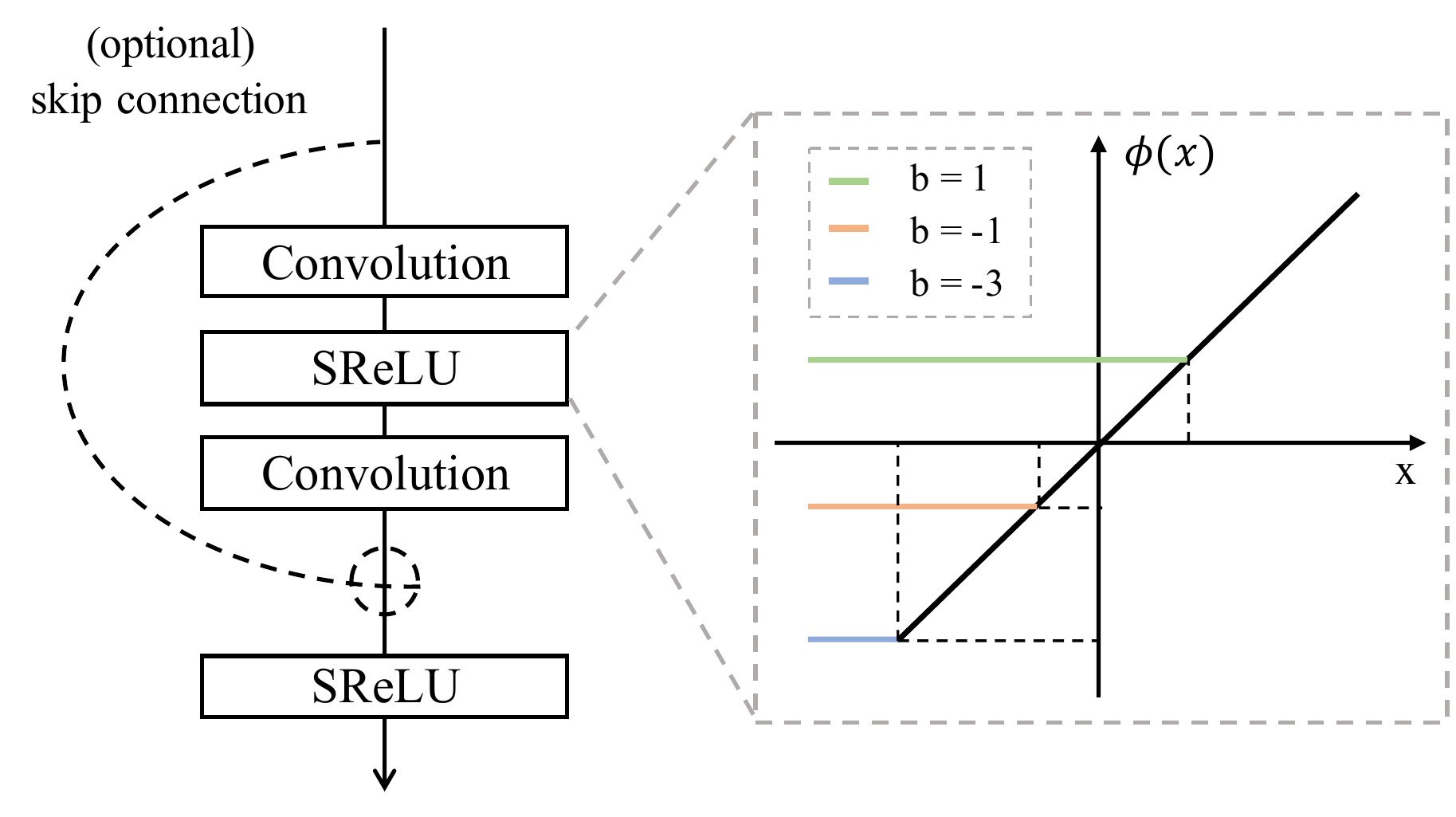}
\label{fig:net}
\vspace{-1.2em}
\caption{A basic block of an Isometric Network (ISONet). Each block contains only convolution and nonlinear activation layers, with an identity skip connections (for R-ISONet only). The convolution is initialized as the (Kronecker) delta kernel and regulated to be (near) orthogonal during training. The activation is Shifted ReLU (SReLU) $\phi(\cdot)$ with a learnable parameter $b$ for obtaining a balance between nonlinearity and isometry. The figure shows three examples of the SReLU with $b = 1, -1$ and $-3$.
}
\end{figure}

Training \emph{deep} ConvNets is inherently difficult~\cite{pascanu2013difficulty,glorot2010understanding}. 
Despite widespread interests in this problem, it has not been made possible until the development of proper weight initialization~\cite{glorot2010understanding,he2015delving}, feature map normalization~\cite{ioffe2015batch},
and residual learning techniques~\cite{he2016deep,srivastava2015highway}. 
Following these initial works, a wide variety of network architectural components including novel nonlinear activation, weight initialization, weight regularization, and residual learning techniques have been purposed in the literature (see Section~\ref{sec:previouswork} for an overview). 
Such techniques are motivated and justified from diverse perspectives, such as prevention of dead neurons \cite{maas2013rectifier}, promotion of self-normalization \cite{klambauer2017self}, reduction of filter redundancy \cite{wang2019orthogonal}, improvement of generalization~\cite{jia2019orthogonal}, to name a few. 

The abundance of existing architectural components and the diversity of their design principles posit ConvNets architectural design a difficult task.
After all, which combination of components should one use for their networks? 
Such a challenge motivates us to pose the following question:
\emph{Is there a central guiding principle for training very deep ConvNets?}

\textbf{Isometric learning.}
We show that the \emph{isometric} property, where each layer of the network preserves the inner product for both forward and backward propagation, plays a central role in training deep ConvNets. 
To illustrate this, we take a minimalist approach and show that a vanilla deep network (i.e., composed of interleaved convolution and nonlinear layers only) can be trained as long as both the convolution and nonlinear activation are close to an isometry. 

Specifically, we design the Isometric Networks (ISONets) where the convolution layer is simply initialized as {\em the identity} and is regularized to be {\em near orthogonal} during training, and the nonlinear layer is the {\em Shifted ReLU} (SReLU) obtained by shifting regular ReLU towards being an identity around the origin with a learnable parameter (see Figure~\ref{fig:net}).
We perform extensive experiments with so designed simple networks on image classification using the ImageNet dataset~\cite{deng2009imagenet,russakovsky2015imagenet}. 
Our results show that an ISONet with more than $100$ layers is trainable and can achieve surprisingly competitive performance. 
To the best of our knowledge, this is the best performing \emph{vanilla} ConvNet for such tasks.

From a practical perspective, isometric learning provides an important guiding principle for addressing challenges in network architectural design. 
One such challenge is that the commonly used normalization layers in modern networks
\cite{ioffe2015batch,ba2016layer,wu2018group}
require certain statistical independence assumptions to hold and large enough batch size or channel number for precise estimation of such statistics.
This drawback significantly limits their applications to robust learning \cite{sun2019test}, contrastive learning \cite{chen2020simple}, implicit models \cite{bai2020multiscale}, object detection and so on. 

To address such a challenge, we develop a Residual ISONet (R-ISONet) that does not contain any normalization layer. 
In particular, 
we introduce skip connection to ISONet as it helps to better promote isometry of the network~\cite{tarnowski2019dynamical}. 
We evaluate the performance of R-ISONet for object detection and instance segmentation on the COCO dataset~\cite{lin2014microsoft}.
In these applications, the batch size is very small due to the high resolution of the images, so that batch normalization becomes ineffective. 
Our experiment shows that R-ISONet obtains better performance than standard ResNet.  
Comparing with existing techniques for addressing the small-batch challenge~\cite{luo2018differentiable,wu2018group,singh2019filter}, R-ISONet has the benefit that it does not suffer from a slowdown during inference. 

Isometry and many closely related notions have been implicitly and explicitly studied and utilized in many previous works for improving deep networks. In particular, techniques to promote isometric property have been widely practiced in the literature. However, they are often used together with many other design techniques and none of the previous work has clearly demonstrated that the isometry property {\em alone} ensures surprisingly strong performance for deep networks. At the end (Section \ref{sec:previouswork}), we discuss how isometric learning offers a simple but unified framework for understanding numerous seemingly different and diverse ideas in the literature, supporting the hypothesis that isometric learning offers a central guiding principle for designing and learning deep networks. 

\section{Isometric Networks}
\label{sec:isonet}

In this section, we introduce the isometric principle and present the isometric network (ISONet).
In Section~\ref{sec:prelim}, we formally introduce the isometric principle in a vanilla net as enforcing both the convolution and nonlinear activation layers to be close to an \emph{isometry}. 
Subsequently, Section~\ref{sec:convolution} derives the notion of isometry for convolution, and explains how it can be enforced at initialization and in the training process.
In Section~\ref{sec:activation}, we discuss principles for designing nonlinear activation functions in isometric learning and argue that one needs to strike a balance between isometry and nonlinearity, which cannot be achieved at the same time. 
Finally, in Section~\ref{sec:residual}, skip connections  (or residual structure) can be naturally introduced to ISONet to further improve isometry, which leads to the R-ISONet.

\subsection{Isometric Learning}
\label{sec:prelim}

We first develop a vanilla network that is composed of interleaved convolution\footnote{We consider convolution for 2D signals (e.g., images), though our analysis trivially generalizes to arbitrary dimensional signals.} and nonlinear activation layers, 
i.e.,
\begin{equation}
    \x^\ell = \phi(\y^\ell), ~~\y^\ell = \cA^\ell \x^{\ell-1}, ~~~\ell=1, \ldots, L,\label{eq:forward}
\end{equation}
where $\cA^\ell: \Re^{C^{\ell-1} \times H\times W} \to \Re^{C^{\ell} \times H\times W}$ denotes a convolution operator, $\phi(\cdot)$ denotes a point-wise nonlinear activation function, and $\x^0 \in \Re^{C^0 \times H \times W}$ is the input signal.
Assuming a squared loss given by Loss$=\frac{1}{2}\|\z - \x^L\|_2^2$, the backward propagation dynamic at $\x^0$ is given by
\begin{equation}\label{eq:backward}
    \frac{\partial \text{Loss}}{\partial \x^0} = (\cA^1)^* \cD^1 \cdots (\cA^L)^* \cD^L (\z - \x^L),
\end{equation}
where $(\cA^\ell)^*: \Re^{ C^{\ell} \times H\times W} \to \Re^{C^{\ell-1} \times H \times W}$ is the adjoint operator of $\cA^\ell$, and $\cD^\ell: \Re^{C^{\ell} \times H \times W} \to \Re^{C^{\ell} \times H \times W}$ operates point-wise by multiplying the $(c, h, w)$-th entry of the input by $\phi'(\y_{c,h,w}^\ell)$. 
For the operator $\cA^\ell$ and $\phi(\cdot)$ in the forward dynamic and $(\cA^\ell)^*$ and $\cD^\ell$ in the backward dynamic, we may define the notion of \emph{isometry} as follows:
\begin{definition}[Isometry]
\label{def:isometry}
A map $\cA: \Re^C \to \Re^M$ is called an isometry if 
\begin{equation}
    \langle \cA \x, \cA \x' \rangle = \langle \x, \x' \rangle, ~~\forall \{\x, \x'\} \subseteq \Re^C.
\end{equation}
\end{definition}

Our ISONet is designed to maintain that all transformations ($\cA, \cA^*, \phi$, and $\cD$) in the forward and the backward dynamics are close to an \emph{isometry}.
In the following, we develop techniques for enforcing isometry in convolution and nonlinear activation layers. 
As we show in Section~\ref{sec:experiments}, combining these components enables effective training of ISONet with more than $100$ layers on image classification tasks.

\subsection{Isometry in Convolution}
\label{sec:convolution}

In this section, we show how to impose the property of isometry for convolutional layers in ConvNets, and explain how it can be achieved at initialization and through network training.

\textbf{Notation.} 
For computation concerning convolution, we treat a 2D signal $\xm \in \Re^{H \times W}$ as a function defined on the discrete domain $[1, \ldots, H] \times [1, \ldots, W]$ and further extended to the domain $\Z \times \Z$ by padding zeros.
Similarly, we regard any 2D kernel $\am \in \Re^{k\times k}$ with $k = 2k_0 + 1$ for some integer $k_0$ as a function defined on the discrete domain $[k_0-1, \ldots, 0, \ldots, k_0+1] \times [k_0-1, \ldots, 0, \ldots, k_0+1]$ and extended to $\Z \times \Z$ by padding zeros. 
$\xm[i, j]$ represents the value of the function $\xm$ at the coordinate $(i, j)$.

Given any $\xm \in \Re^{H \times W}$ and $\am \in \Re^{k \times k}$, the convolution of $\xm$ and $\am$ is defined as
\begin{equation}\label{eq:def-conv}
    (\am * \xm)[i, j] = \sum_{p=-k_0}^{k_0} \sum_{q=-k_0}^{k_0} \xm[i-p, j-q] \cdot \am[p, q], 
\end{equation}
and the correlation of $\xm$ and $\am$ is defined as
\begin{equation}\label{eq:def-corr}
    (\am \star \xm)[i, j] = \sum_{p=-k_0}^{k_0} \sum_{q=-k_0}^{k_0} \xm[i+p, j+q] \cdot\am[p, q].
\end{equation}

In contrast to the conventions in signal processing, the convolution layers (i.e. $\{\cA^\ell\}_{\ell=1}^L$ in \eqref{eq:forward}) in modern deep learning frameworks actually perform multi-channel \textit{correlation} operations that map a $C$-channel 2D signal to an $M$-channel 2D signal. Let $\x = (\xm_1, \ldots, \xm_C)\in \Re^{C\times H\times W}$ be the input signal where $\xm_c \in \Re^{H \times W}$ for each $c \in \{1, \ldots, C\}$, and let
\begin{equation}\label{eq:kernel}
\resizebox{.9\hsize}{!}{$
    \A =
    \begin{pmatrix}
    \am_{11} & \am_{12} & \am_{13} & \dots & \am_{1C} \\
    \am_{21} & \am_{22} & \am_{23} & \dots & \am_{2C} \\
    \vdots & \vdots & \vdots & \ddots & \vdots \\
    \am_{M1} & \am_{M2} & \am_{M3} & \dots & \am_{MC}
\end{pmatrix}
\in \Re^{M\times C \times k \times k},$
}
\end{equation}
be the convolution kernel where each $\am_{m,c}$ is a kernel of size $k \times k$ for $c\in\{1, \ldots, C\}$ and $m \in \{1, \ldots, M\}$.
The convolution operator $\cA$ associated with $\A$ is given by
\begin{equation}\label{eq:conv_A}
    \cA \x := \sum_{c=1}^C \big(\am_{1c} \star \xm_c, \ldots, \am_{Mc} \star\xm_c\big) \in \Re^{M \times H \times W}.
\end{equation}

The adjoint of $\cA$ that appears in the backward dynamics \eqref{eq:backward}, denoted as $\cA^*$, is a mapping from $\Re^{M \times H\times W}$ to $\Re^{C\times H \times W}$:
\begin{equation}\label{eq:conv_At}
    \cA^* \y := \sum_{m=1}^M \big(\am_{m1} * \ym_m, \ldots, \am_{mC} * \ym_m\big) \in \Re^{C \times H \times W},
\end{equation}
where
$\y = (\ym_1, \ldots, \ym_M)\in \Re^{M \times H \times W}$ and $\ym_m \in \Re^{H \times W}$.

We first study under what conditions of kernel $\A$ such that $\cA$ and $\cA^*$ are isometric. Starting from Definition~\ref{def:isometry} we arrive at the following theorem (see appendix for the proof).

\begin{theorem}\label{thm:convolution_isometry}
Given a convolution kernel $\A\in \Re^{M\times C \times k \times k}$ in \eqref{eq:kernel}, the operator $\cA$ is an isometry if and only if
\begin{equation}\label{eq:isometry-conv}
    \sum_{m=1}^M \am_{mc} \star \am_{mc'} = 
    \begin{cases}
        \mathbf{\delta} & ~\text{if}~c = c',\\
        \0 & ~\text{otherwise},\\
    \end{cases}
\end{equation}
and the operator $\cA^*$ is an isometry if and only if
\begin{equation}\label{eq:coisometry-conv}
    \sum_{c=1}^C \am_{mc} \star \am_{m'c} = 
    \begin{cases}
    \mathbf{\delta} & ~\text{if}~m = m',\\
    \0 & ~\text{otherwise}.\\
    \end{cases}
\end{equation}
In above, $\mathbf{\delta}$ is the Kronecker delta function defined on $\Z \times \Z$ that takes value $1$ at coordinate $(0, 0)$ and $0$ otherwise.
\end{theorem}

For the case of $M = C$, the isometry of $\cA$ is equivalent to the isometry of $\cA^*$\footnote{When $M=C$, the isometry of $\cA$ implies that it is surjective, therefore unitary, hence $\cA^*$ is also an isometry.}.
We refer to $\A$ as an \emph{orthogonal convolution} kernel if $M=C$ and either $\cA$ or $\cA^*$ is isometric. 

On the other hand, if $M \ne C$ then $\cA$ and $\cA^*$ cannot be both isometric. 
Nonetheless, we can still enforce isometry of $\cA$ (in the case $M\ge C$) or the isometry of $\cA^*$ (in the case $M\le C$) by the conditions \eqref{eq:isometry-conv} and \eqref{eq:coisometry-conv}, respectively. 

\vspace{1em}
\begin{example}[Delta kernel]
\label{eg:delta}
    We refer to a convolution kernel $\A \in \Re^{ C \times M \times k \times k}$ as the (Kronecker) delta kernel, denoted as $\mathbf{\delta}^{C\times M \times k \times k}$, if its entries indexed by $(i, i, k_0, k_0)$ (assuming $k = 2k_0+1$) for all $i \in \{1, \ldots, \min(C, M)\}$ are set to $1$ and all other entries are set to zero.
    If $M \ge C$ (resp., $M \le C$), then the operator $\cA$ (resp., $\cA^*$) associated with a delta kernel is an isometry.
    Moreover, if $M = C$ then a delta kernel is orthogonal.
\end{example}

\textbf{Isometry at initialization.} As shown in Example~\ref{eg:delta}, we may initialize all convolution kernels to be isometric by setting them to be the delta kernel, which we refer to as the Delta initialization. 
Other orthogonal initialization, such as the Delta Orthogonal initialization \cite{xiao2018dynamical}, are also plausible choices. We empirically find that Delta initialization works well and often outperforms Delta Orthogonal initialization. Such initialization is also commonly used in initializing Recurrent Neural Networks~\cite{le2015simple}.

\textbf{Isometry during training.} Isometry at initialization does not guarantee that isometry will be preserved through the training process. 
In addition to Delta initialization, we enforce isometry by penalizing the difference between the left and right hand sides of \eqref{eq:isometry-conv} (or \eqref{eq:coisometry-conv}).
As we shown below, this can be easily implemented via modern deep learning packages.

In particular, given an input that contains a batch of $N$ multi-channel signals $\X := (\x_1, \ldots, \x_N) \in \Re^{N \times C \times H \times W}$ where each $\x_n \in \Re^{C \times H \times W}$ for each $n \in \{1, \ldots, N\}$, and a convolution kernel $\A \in \Re^{M\times C \times k \times k}$ defined as in \eqref{eq:kernel}, a convolution function in a typical deep learning framework, denoted as $\text{Conv}(\A, \X)$, performs correlation of $\A$ on each of the $N$ signals and stack the resulting $N$ multi-channel signals to a tensor:
\begin{equation}\label{eq:conv-operator}
    \text{Conv}(\A, \X) = (\cA \x_1, \ldots, \cA \x_N).
\end{equation}
By replacing $\X$ in \eqref{eq:conv-operator} with the kernel $\A$ itself and using \eqref{eq:conv_A}, we obtain
\begin{equation}
\begin{split}\label{eq:convAA}
    & \text{Conv}(\A, \A) \\
    =& \; \big( \cA(\am_{11}, \ldots, \am_{1C}), \ldots, \cA(\am_{M1}, \ldots, \am_{MC}) \big)\\
    =& 
    \begin{small}
    \begin{pmatrix}
        \sum_{c=1}^C \am_{1c}\star \am_{1c} & \cdots & \sum_{c=1}^C \am_{Mc}\star \am_{1c}\\
        \vdots & \ddots & \vdots \\
        \sum_{c=1}^C \am_{1c}\star \am_{Mc} & \cdots & \sum_{c=1}^C \am_{Mc}\star \am_{Mc}
    \end{pmatrix}.
    \end{small}
\end{split}
\end{equation}
Comparing \eqref{eq:convAA} with the condition in \eqref{eq:coisometry-conv}, we see that the operator $\cA^*$ is an isometry if and only if Conv$(\A, \A)$ is a delta kernel $\mathbf{\delta}^{M \times M \times k \times k}$. 
Similarly, $\cA$ is an isometry if and only if Conv$(\A^\top, \A^\top) = \mathbf{\delta}^{C \times C \times k \times k}$, where $\A^\top$ denotes a kernel with the first two dimension of $\A$ transposed. 
From these results, we can enforce isometry by adding one of the following regularization terms to the objective function:
\begin{align}\label{eq:orth-reg}
    L(\A) &= \frac{\gamma}{2} \|\text{Conv}(\A, \A) - \mathbf{\delta}^{M \times M \times k \times k}\|_F^2, ~~\text{or}\\
    L(\A^\top) &= \frac{\gamma}{2} \|\text{Conv}(\A^\top, \A^\top) - \mathbf{\delta}^{C \times C \times k \times k}\|_F^2,
\end{align}
where $\gamma$ is a regularization coefficient. We use $L(\A)$ when $C > M$ and $L(\A^\top)$ otherwise.

\subsection{Isometry in Nonlinear Activation}
\label{sec:activation}

The rectified linear unit (ReLU) is one of the most popular activation functions for deep learning applications \cite{nair2010rectified}. 
Defined entry-wise on the input as $\phi(x) = \max(0, x)$, ReLU is an identity map for nonnegative input, but completely removes all negative component.
For an input signal that is normalized to zero mean and unit variance, ReLU is far from being an isometry. 
How can we develop an isometric nonlinear activation layer?

\begin{figure}[t]
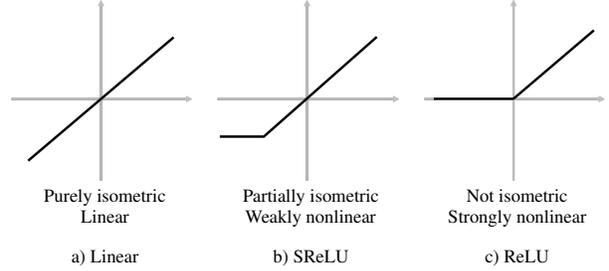

\centering
    \begin{minipage}[t]{0.3\linewidth}
    \centering
    \includegraphics[width=\linewidth]{figs/srelu/linear.pdf}
    \scriptsize{Purely isometric \\ Linear \\ \vspace{1em} a) Linear}
    \end{minipage}
    ~
    \begin{minipage}[t]{0.3\linewidth}
    \centering
    \includegraphics[width=\linewidth]{figs/srelu/srelu.pdf}
    \scriptsize{Partially isometric \\ Weakly nonlinear \\ \vspace{1em} b) SReLU}
    \end{minipage}
    ~
    \begin{minipage}[t]{0.3\linewidth}
    \centering
    \includegraphics[width=\linewidth]{figs/srelu/relu.pdf}
    \scriptsize{Not isometric \\ Strongly nonlinear \\ \vspace{1em} c) ReLU}
    \end{minipage}
    \caption{Illustration of isometry in nonlinear activation}
    \label{fig:srelu}
\end{figure}

Unfortunately, isometry is innately at odds with nonlinearity. 
By Mazur-Ulam theorem, any surjective isometry between two normed spaces over $\Re$ is linear up to a translation \cite{nica2013mazur}. 
Therefore, isometry can never be achieved if any nonlinearity is present. On the other hand, a purely isometric network without nonlinearity is not interesting either, since the entire network becomes a linear transformation which has very limited learning ability.
The design of the nonlinear activation function requires striking a good {\em trade-off between isometry and nonlinearity}.

We advocate using a Shifted ReLU (SReLU) for achieving such a trade-off. SReLU is obtained by interpolating regular ReLU (for obtaining nonlinearity) and the identity (for obtaining isometry). 
Given a signal $\y \in \Re^{C\times H \times W}$ where $C$ is the number of channels, SReLU is defined point-wise as
\begin{equation}\label{eq:SReLU}
    \phi_{b}(y) = \max(y, b),
\end{equation}
where $b$ is a parameter that is shared for each channel of $\y$.
SReLU becomes the regular ReLU if $b = 0$ and the identity if $b \to -\infty$, therefore can be interpreted as a trade-off between nonlinearity and isometry  (see Figure~\ref{fig:srelu}). 
In terms of backward dynamic (see \eqref{eq:backward}), the operator $\cD$ associated with $\phi_{b}(\cdot)$ is given as $\cD(x) = 1_{y \ge b} \cdot x$, which is an identity map if $y$ is larger than $b$.

In our experiments, we find it beneficial to initialize $b$ to be negative values, which indicates the importance of such a trade-off between ReLU and identity. 
For simplicity, in all our experiments we initialize $b$ to be $-1$ and optimize it in the training process. 

\subsection{Isometry in Residual Structure}
\label{sec:residual}

We may further improve the isometric property of each layer of ISONet with a residual component \cite{he2016deep}, which we call a residual ISONet (R-ISONet).
The core idea is to introduce a skip connection to ISONet, so that the network learns a \emph{residual} component that is added onto the path of signal propagation (see Figure~\ref{fig:net}).
Such a network architecture is automatically near isometric if the residual component is small enough relative to the identity shortcut~\cite{tarnowski2019dynamical}.
Motivated by such an observation, we add a scalar multiplier $s$ at the end of each residual branch and initialize $s$ to be zero ($s$ is shared within each channel).  
Similar design ideas have previously been explored to improve ResNet~\cite{szegedy2017inception,goyal2017accurate, zhang2019fixup}. 

Since skip connection helps to enforce isometry, R-ISONet exhibits better performance than ISONet. 
For image classification on ImageNet, R-ISONet obtains almost as good performance as regular ResNet, even though R-ISONet has no normalization layers.  
Since no normalization layer is required, R-ISONet is particularly suited for applications where normalization layers are ineffective. 
As we show in the next section, R-ISONet obtains better performance than regular ResNet for object detection tasks on COCO dataset. 

\section{Experiments}
\label{sec:experiments}

\subsection{Experimental Setup}

\textbf{Setup.}
We test the performance of ISONets on the ILSVRC-2012 image classification dataset~\cite{deng2009imagenet, russakovsky2015imagenet}. The training set contains $\app1.28$ million images from $1,\!000$ categories. Evaluation is performed on the validation set which contains $50,\!000$ images. 1-Crop, Top-1 accuracy is reported.

\textbf{Network architectures.} 
The design of network architecture follows from the design of ResNet \cite{he2016deep, goyal2017accurate}, with the only difference being that we change the residual blocks to the block shown in Figure~\ref{fig:net}. In particular, the number of channels and layers in each stage remain the same as ResNet. We find that (R-)ISONet tends to overfit since BatchNorm is removed. Therefore we add one dropout layer right before the final classifier. The dropout probability is $0.4$ for R-ISONet and $0.1$ for ISONet.

\textbf{Implementation details.} 
The hyperparameter settings follow from prior works~\cite{he2016deep, goyal2017accurate}. All models are trained using SGD with weight decay $0.0001$, momentum $0.9$ and mini-batch size $256$. The initial learning rate is 0.1 for 
R-ISONet and 0.02 for ISONet. The models are trained for 100 epochs with the learning rate subsequently divided by a factor of $10$ at the $30$th, $60$th and $90$th epochs. To stabilize training in the early stage, we perform linear scheduled warmup~\cite{goyal2017accurate} for the first $5$ epochs for both our methods and baseline methods. The orthogonal regularization coefficient $\gamma$ in \eqref{eq:orth-reg} (when used) is 0.0001 except for ISONet-101 where a stronger regularization (0.0003) is needed. For other baselines without normalization or skip connections, we choose the largest learning rate that makes them converge.

\subsection{ISONet: Isometry Enables Training Vanilla Nets}

In this section, we verify through extensive experiments that isometry is a central principle for training deep networks. 
In particular, we show that more than $100$-layer ISONet can be effectively trained with neither normalization layers nor skip connections.
In addition, we demonstrate through ablation study that all of the isometric components in ISONet are necessary for obtaining good performance.

\begin{table}[t]
    \centering
    \small
    \setlength{\tabcolsep}{4pt}
    \renewcommand{\arraystretch}{1.5}
    \begin{tabular}{c|c|ccc|c}
    & Method & SReLU & {\begin{mytabular}[1.2]{c} Delta \\ Init. \end{mytabular}} & \begin{mytabular}[1.2]{c} Ortho. \\ Reg. \end{mytabular} & \begin{mytabular}[1.2]{c} Top-1 \\ Acc. (\%) \end{mytabular}  \\
    \hline
    \hline
    (a) & \demph{ResNet} & & & & \demph{73.29}\\
    \hline
    (b) & Vanilla & & & & 63.09 \\
    \hline
    (c) & & & \checkmark & \checkmark & 46.83 \\
    (d) & & \checkmark & & & 67.35 \\
    (e) & & \checkmark & & \checkmark & 68.50 \\
    (f) & & \checkmark & \checkmark & & 68.55 \\ 
    \hline
    (g) & ISONet & \checkmark & \checkmark & \checkmark & \textbf{70.45} \\
    \end{tabular}
    \caption{\textbf{Isometric learning (with SReLU, Delta initialization and orthogonal regularization) enables training ISONets on ImageNet without BatchNorm and skip connection. }
    (a) Regular $34$-layer ResNet with ReLU activation, Kaiming initialization and no orthogonal regularization. (b) Same as ResNet but without BatchNorm and skip connection. (g) Our ISONet with the same backbone as Vanilla. (c-f) Ablation for (g) that shows that the combination of all three isometric learning components is necessary for effectively training ISONet.
    }
    \label{tab:abl_iso}
\end{table}

\begin{figure*}[t]
\centering  
\subfigure{\includegraphics[width=0.24\linewidth]{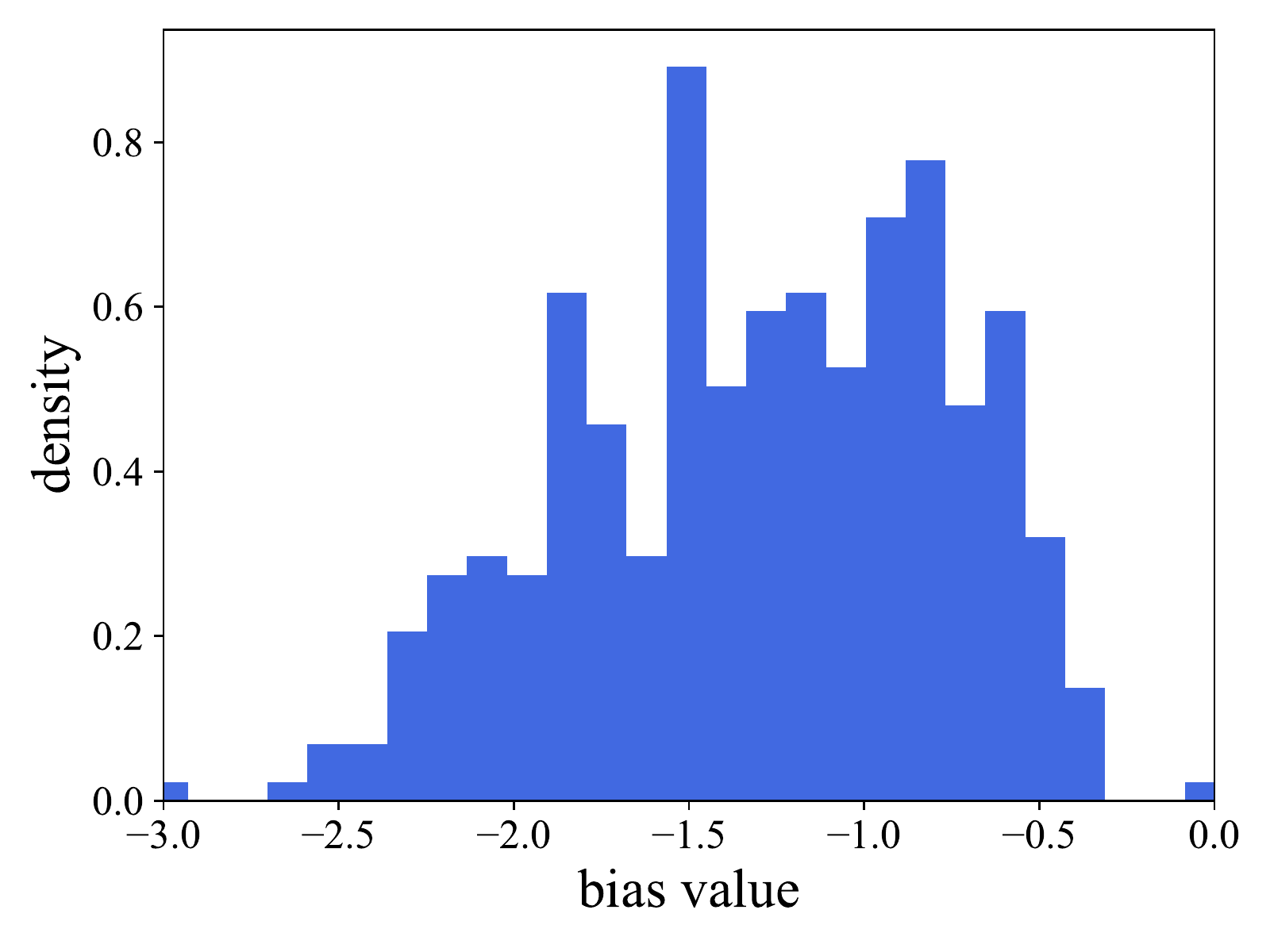}}
\subfigure{\includegraphics[width=0.24\linewidth]{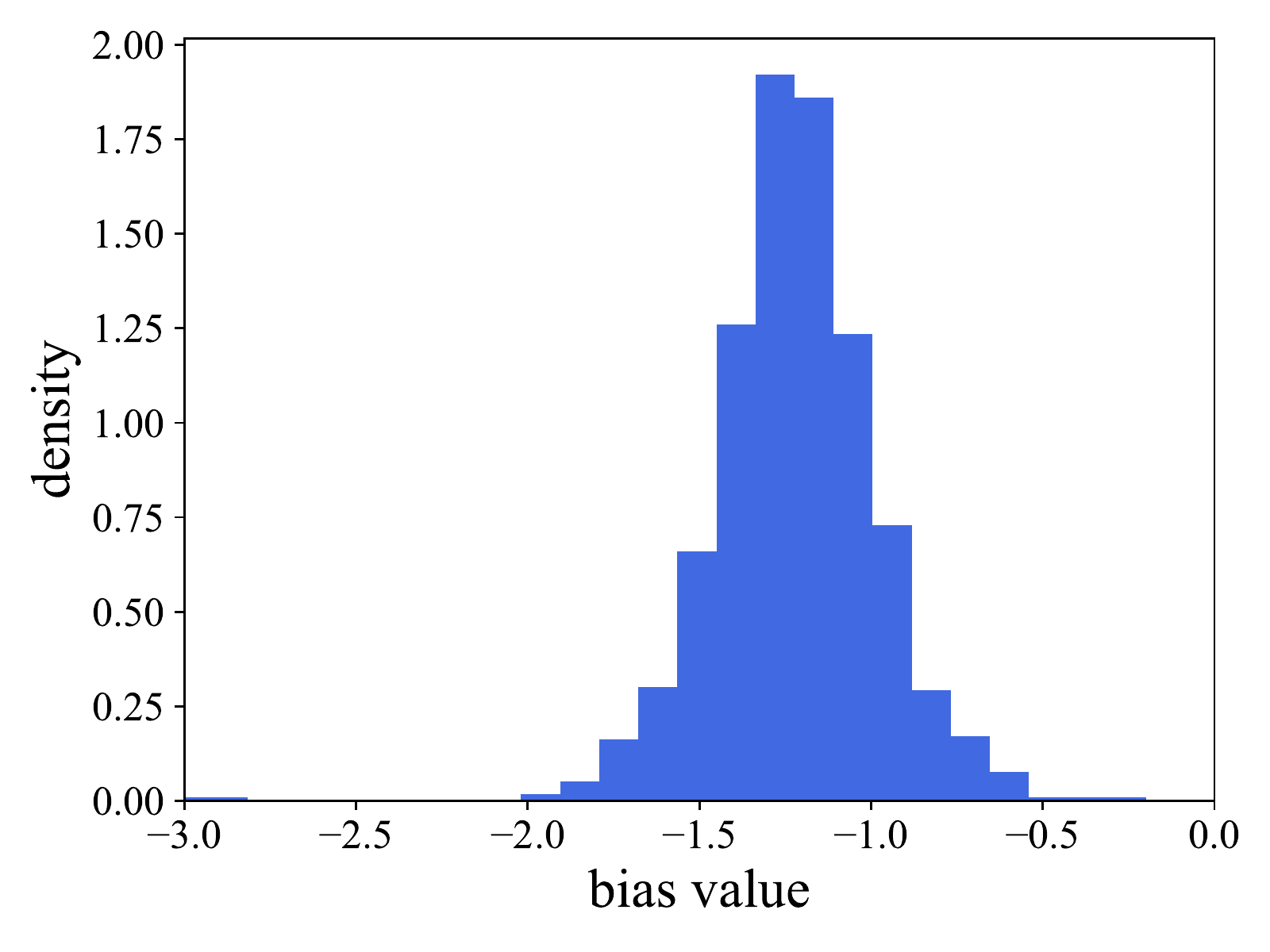}}
\subfigure{\includegraphics[width=0.24\linewidth]{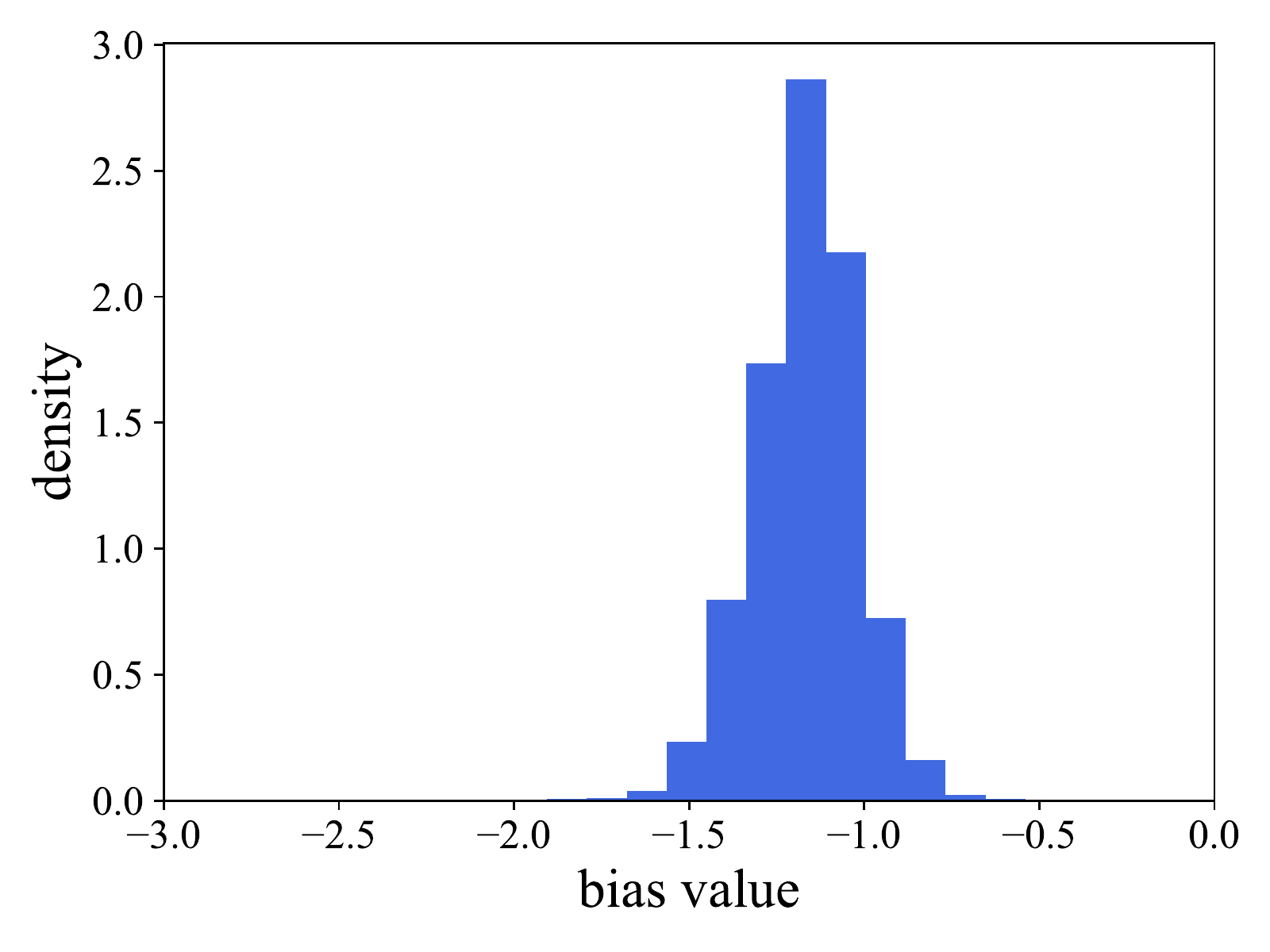}}
\subfigure{\includegraphics[width=0.24\linewidth]{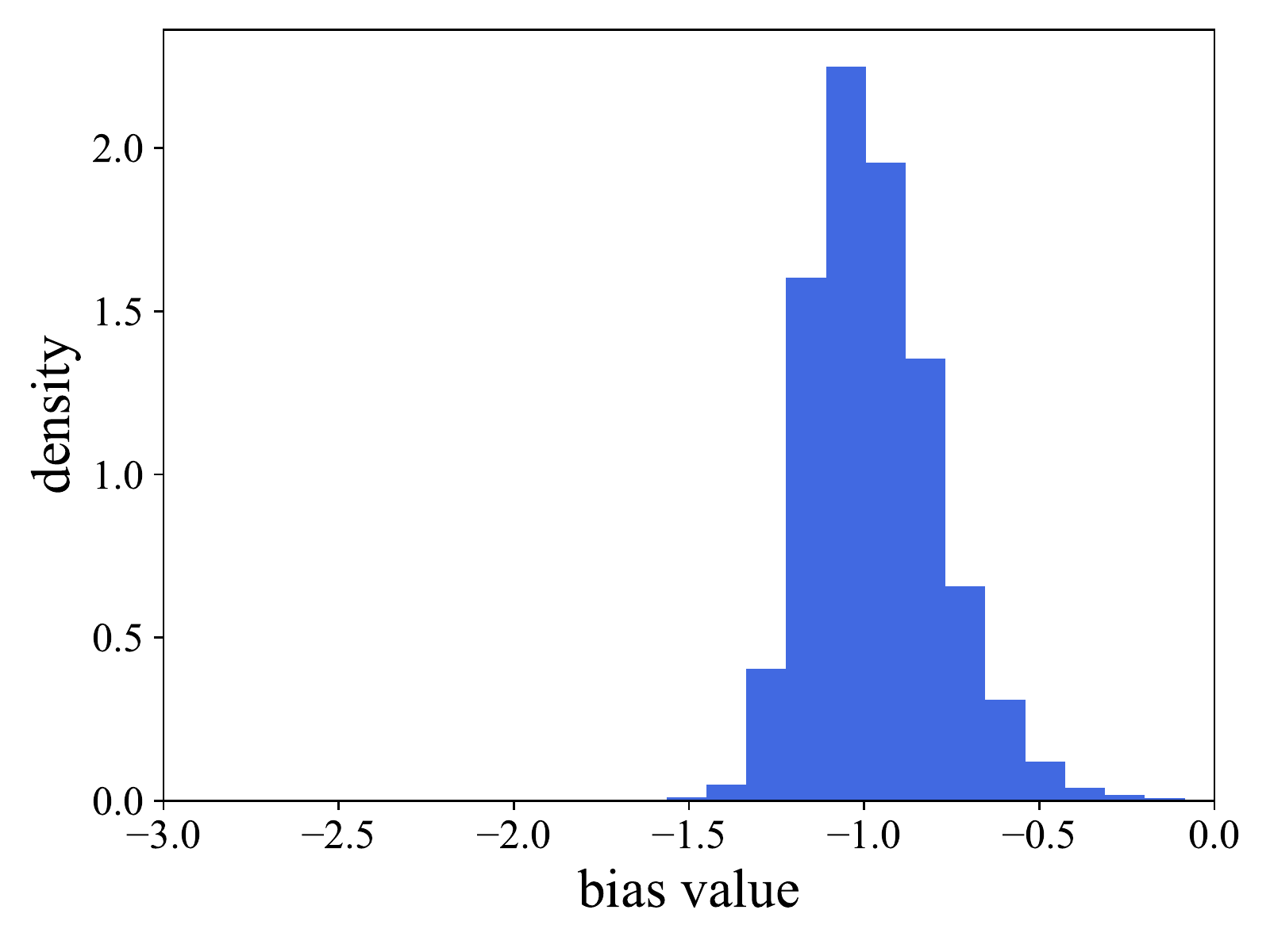}}
\vspace{-8pt}
\caption{Histogram of the parameter $b$ in SReLU computed for each of the four stages of a 34-layer ISONet.
}\label{fig:bias}
\end{figure*}

\begin{table}[t]
    \centering
    \small
    \setlength{\tabcolsep}{3.5pt}
    \renewcommand{\arraystretch}{1.6}
    \begin{tabular}{c|ccccc}
    Orthogonal coefficient $\gamma$ & 0 & $1e^{-5}$ & $3e^{-5}$ & $1e^{-4}$ & $3e^{-4}$ \\
    \hline
    \hline
    Top-1 Accuracy (\%) & 68.55 & 70.08 & 70.44 & \textbf{70.45} & 69.32 \\
    \end{tabular}
    \caption{\textbf{Effect of orthogonal coefficients for 34-layer ISONets}. The performance is not sensitive to the coefficient in a wide range.}
    \label{tab:abl_coeff}
    \vspace{-0.5em}
\end{table}

\begin{table}[t]
    \centering
    \small
    \setlength{\tabcolsep}{4pt}
    \renewcommand{\arraystretch}{1.4}
    \begin{tabular}{c|c|c|c|c}
    & Activation & Adaptive $b$? & ~~Initial $b$~~ & Top-1 Acc. (\%) \\
    \hline
    \hline
    (a) & ReLU & & - & 46.83 \\
    \hline
    (b) & \multirow{3}{*}{SReLU} & & -1.0 & 70.27 \\
    (c) & & \checkmark & -1.0 & 70.45 \\
    (d) & & \checkmark & -2.0 & 70.04 \\
    \end{tabular}
    \caption{\textbf{Effect of parameter $b$ in SReLU for 34-layer ISONets. }}
    \label{tab:abl_act}
    \vspace{-0.5em}
\end{table}

\textbf{Ablation study of ISONet components.}
To demonstrate how each isometric component affects the trainability of deep neural networks, we perform experiments with $34$-layer vanilla neural networks and report results in Table~\ref{tab:abl_iso}.
In Table~\ref{tab:abl_iso} (a), we show the result of regular ResNet.
If the normalization layers and skip connections in ResNet are removed, the network performance drops by more than $10$ points as shown in Table~\ref{tab:abl_iso} (b). 
This shows a plain vanilla network cannot be easily optimized.
In fact, $34$-layer is the deepest vanilla network we found trainable in our experiments.
In contrast, ISONet with neither normalization layers nor skip connections can be effectively trained and obtain competitive $70.45\%$ Top-1 accuracy, as shown in Table~\ref{tab:abl_iso} (g). 
To our best knowledge, this is the only vanilla network that achieves $>70\%$ accuracy on ImageNet.

We further show that all the three isometric components (i.e., Delta initialization, orthogonal regularization and SReLU) are indispensable for training deep ISONet.
Firstly, enforcing isometry in convolution alone produces \app$46\%$ Top-1 Accuracy, which is far lower than ISONet (see Table~\ref{tab:abl_iso} (c)). 
The reason is that orthogonal convolution is not suitable for vanilla networks with ReLU activation functions, since the energy may decrease exponentially through multiple layers of the network.
Secondly, enforcing isometry in activation functions alone cannot achieve good performance, as shown in Table~\ref{tab:abl_iso} (d). 
Thirdly, for convolution layers, isometry must be enforced both at initialization as well as during training. 
This can be seen in Table~\ref{tab:abl_iso} (e) and (f), where the performance of ISONet drops by $2$ points if either Delta initialization or orthogonal regularization is removed.
This demonstrates the critical role of isometry during the entire training process. 

\textbf{Analysis for orthogonal regularization.} 
To understand the effect of the orthogonal regularization term in \eqref{eq:orth-reg}, we conduct experiments with different values of the regularization coefficient $\gamma$ and report Top-1 accuracy in Table~\ref{tab:abl_coeff}. 
It is shown that our method is not sensitive to the choice of $\gamma$ in a wide range (from $1e^{-5}$ to $1e^{-4}$), whereas too large ($3e^{-4}$) or no regularization will hurt the performance.

\textbf{Analysis for SReLU.} 
To verify the design idea of SReLU, we perform an ablation study on the effect of parameter $b$ in \eqref{eq:SReLU} and report the results in Table~\ref{tab:abl_act}. 
Comparing (b) and (c), a trainable threshold in SReLU is better than a fixed one.
Comparing (c) and (d), changing the initial value of $b$ from $-1$ to $-2$ reduces the accuracy by less than $0.5$ percent.
We also plot the histogram of the learned $b$ in SReLU computed within each of the four stages (corresponding to spatial resolutions $\{56, 28, 14, 7\}$) of ISONet and report the results in Figure~\ref{fig:bias}. 
In all cases, the histogram is bell-shaped and concentrates near $-1$. This explains why a fixed $b= -1$ produces a good performance as shown in Table~\ref{tab:abl_act} (b).

\textbf{Training very deep ISONets. }
To demonstrate the effectiveness of isometric principle for training deep networks, we report performance of ISONet with increasing number of layers and report the results in
Table~\ref{tab:depth}.
Firstly, we observe that a standard vanilla network, either without (Table~\ref{tab:depth} (c)) or with (Table~\ref{tab:depth} (d)) BatchNorm, cannot converge when networks are extremely deep. 
In contrast, ISONet maintains a competitive accuracy when the depth is beyond $100$ layers. 
Adding dropout to ISONet further improves the performance by \app 0.5 points. On the other hand, the performance ISONet is not as good as ResNet.
In the following, we show that when combined with skip connection, the performance of our R-ISONet is on par with ResNet.

\begin{table}[t]
    \centering
    \small
    \setlength{\tabcolsep}{4pt}
    \renewcommand{\arraystretch}{1.5}
    \begin{tabular}{c|c|c|cccc}
    & \multirow{2}{*}{Method} & \multirow{2}{*}{\footnotesize{Dropout}} & \multicolumn{4}{|c}{Top-1 Accuracy (\%)} \\
    & & & d=18 & d=34 & d=50 & d=101 \\
    \hline
    \hline
    (a) & \multirow{2}{*}{\demph{ResNet}} &  & \demph{69.67} & \demph{73.29} & \demph{76.60} & \demph{77.37} \\
    (b) & & \demph{\checkmark} & \demph{68.91} & \demph{73.35} & \demph{76.40} & \demph{77.99} \\
    \hline
    (c) & Vanilla & & 65.67 & 63.09 & \footnotesize{N/A} & \footnotesize{N/A} \\
    (d) & Vanilla+BN & & 68.98 & 69.43 & 70.00 & \footnotesize{N/A} \\
    \hline
    (e) & \multirow{2}{*}{ISONet} & & 67.94 & 70.45 & 70.73 & 70.38 \\
    (f) & & \checkmark & 68.10 & 70.90 & 71.20 & 71.01
    \end{tabular}
    \caption{\centering\textbf{ISONet with varying depth $d \in \{18, 34, 50, 101\}$ can be effectively trained on ImageNet.} (a, b) Regular ResNet with or without dropout. 
    (c) Same as ResNet but without BatchNorm and skip connection. (d) Same as ResNet but without skip connection.
    (e, f) Our ISONet with the same backbone as Vanilla. 
    N/A stands for not converging or obtaining an accuracy lower than $40$.
    }
    \vspace{-0.5em}
    \label{tab:depth}
\end{table}

\begin{table}[h!]
    \centering
    \small
    \setlength{\tabcolsep}{4pt}
    \renewcommand{\arraystretch}{1.5}
    \begin{tabular}{c|c|c|cccc}
    & \multirow{2}{*}{Method} & \multirow{2}{*}{\footnotesize{Dropout}} & \multicolumn{4}{|c}{Top-1 Accuracy (\%)} \\
    & & & d=18 & d=34 & d=50 & d=101 \\
    \hline
    \hline
    (a) & \multirow{2}{*}{ResNet} & & 69.67 & 73.29 & 76.60 & 77.37 \\
    (b) & & \checkmark & 68.91 & 73.35 & 76.40 & 77.99 \\
    \hline
    (c) & R-Vanilla & & 65.66 & \footnotesize{N/A} & \footnotesize{N/A} & \footnotesize{N/A} \\
    \hline
    (d) & Fixup & & 68.63 & 71.28 & 72.40 & 73.18 \\
    (e) & Fixup++ & & 67.37 & 72.56 & 76.00 & 76.17 \\
    \hline
    (f) & \multirow{2}{*}{R-ISONet} & & 69.06 & 72.17 & 74.20 & 75.44 \\
    (g) & & \checkmark & 69.17 & 73.43 & 76.18 & 77.08 
    \end{tabular}
    \caption{\textbf{Performance of R-ISONet with varying depth $d \in \{18, 34, 50, 101\}$ on ImageNet. }
    All networks do not have BatchNorm except (a, b).
    (c) ResNet with all BatchNorm removed. (d, e) Same as (c), but with Fixup initialization \cite{zhang2019fixup}. (f, g) Our R-ISONet performs comparable to ResNet in (a, b) and better than all networks in (c, d, e).  
    N/A stands for not converging or obtaining an accuracy lower than $40$. Fixup results are reproduced from running the code in \cite{zhang2019fixup} with $100$ training epochs. Fixup++ denotes Fixup with Mixup data augmentation.
    }
    \label{tab:depth2}
\end{table}

\subsection{R-ISONet: a Practical Network without Normalization Layers}
\label{sec:exp-R-ISONet}

In this section, we evaluate the performance of R-ISONet for ImageNet classification and show that it is comparable to regular ResNet, despite the fact that R-ISONet does not contain BatchNorm. The results are reported in Table~\ref{tab:depth2}.

From Table~\ref{tab:depth2} (c), ResNet without BatchNorm cannot converge for $d\ge34$. In contrast, our R-ISONet (Table~\ref{tab:depth2} (f)) can be effectively trained for varying network depth.
In addition, R-ISONet is better than previous methods without BatchNorm such as Fixup~\cite{zhang2019fixup} (Table~\ref{tab:depth2} (d)).

Since R-ISONet is prone to overfitting due to the lack of BatchNorm, we add dropout layer right before the final classifier and report the results in Table~\ref{tab:depth2} (g). 
The results show that R-ISONet is comparable to ResNet with dropout (Table~\ref{tab:depth2} (b)) and is better than Fixup with Mixup regularization~\cite{zhang2017mixup}.

The superior performance that R-ISONet obtains even without any normalization layers makes it a favorable choice for tasks where statistics in BatchNorm cannot be precisely estimated.
In Section~\ref{sec:exp-object}, we demonstrate such advantage of R-ISONet for object detection and instance segmentation tasks. 

\subsection{Transfer Learning on Object Detection}
\label{sec:exp-object}
 
We further evaluate our method for object detection and instance segmentation tasks on COCO dataset~\cite{lin2014microsoft}.
The dataset contains $115$k training images and $5$k validation images. We use the mAP (over different IoU threshold) metric to evaluate the performance of the model, and the ``dilated conv5'' variant of Faster RCNN~\cite{ren2015faster} as our detector. 
The backbone of the network includes all the convolution layers in R-ISONet.
The RoIAlign operator is added on top of the last convolution with the last stage having convolution with stride $1$ and dilation $2$. 
Two fully-connected layers are then used to predict the bounding box scores and regression output using this feature.
This design follows prior works~\cite{wu2019detectron2, dai2017deformable, he2019momentum}. 
The hyper-parameter settings of both ResNet and R-ISONet follows detectron2~\cite{wu2019detectron2}. We train our model for 90k iterations. The learning rate is initialized to be $0.01$ and divide by $10$ at 60k and 80k iterations.
The training is performed on 8 GPUs, each of which holds 2 images. To keep a fair comparison with standard protocols in object detection, dropout is not added.

The results are reported in Table~\ref{tab:abl_det}. 
Although the classification accuracy of our R-ISONet is lower than that of ResNet with the same depth, the detection and instance segmentation performance of R-ISONet is better. This demonstrates that our model has better feature transfer abilities and can mitigate the disadvantages introduced by BatchNorm.

\begin{table}[t]
    \centering
    \small
    \setlength{\tabcolsep}{5pt}
    \renewcommand{\arraystretch}{1.5}
    \begin{tabular}{c|c|cc}
    & Methods & mAP$^{\text{bbox}}$ & mAP$^{\text{mask}}$ \\
    \hline
    \hline
    \multirow{2}{*}{34 layer} & ResNet & 35.0 & 32.2 \\
    & R-ISONet & \textbf{36.2} & \textbf{33.0} \\
    \hline
    \multirow{2}{*}{50 layer} & ResNet  & 37.0 & 33.9 \\
    & R-ISONet & \textbf{37.3} & \textbf{34.4} \\
    \end{tabular}
    \caption{\textbf{Performance of R-ISONet with varying depth for object detection on COCO dataset. } R-ISONet \textit{outperforms} standard ResNet, indicating that our model has better transfer ability.
    }
    \label{tab:abl_det}
    \vspace{-0.5em}
\end{table}

\section{Related Work and Discussion}
\label{sec:previouswork}

In this paper, we contend that isometry is a central design principle that enables effective training and learning of deep neural networks. In the literature, numerous instantiations of this principle have been explicitly or implicitly suggested, studied, and exploited, often as an additional heuristic or regularization, for  improving existing network training or performance. Now supported by the strong empirical evidence presented earlier about isometric learning, we reexamine these ideas in the literature from the unified perspective of isometry.

Arguably, the notion of isometry was first explored in the context of weight initialization (Section \ref{sec:c_init}) and then in diverse contexts such as weight regularization (Section~\ref{sec:c_regu}), design of nonlinear activation (Section~\ref{sec:c_activation}), and training techniques for residual networks (Section~\ref{sec:c_residual}).

\subsection{In Context of Weight Initialization}
\label{sec:c_init}

Early works on weight initialization are based on the principle that the variance of the signal maintains a constant level as it propagates forward or backward through multiple layers \cite{lecun2012efficient,glorot2010understanding}. 
A popular method that provides such a guarantee is the Kaiming Initialization \cite{he2015delving} which derives a proper scaled Gaussian initialization for weights in vanilla convolutional networks with ReLU activation functions.
Derivations for general activation functions beyond ReLU is more difficult, but nonetheless can be achieved by working in the regime where the network is infinitely wide using mean-field theory \cite{poole2016exponential,schoenholz2016deep}. 
From the perspective of isometry, the aforementioned works guarantee that the average of the squared singular values of the input-output Jacobian matrix is close to $1$. 
However, this \emph{does not mean that the Jacobian is an isometry}, which requires that \emph{all} the singular values concentrate at $1$.
In fact, with the common practice of Gaussian weight initialization, isometry can never be achieved regardless of the choice of activation functions \cite{pennington2018emergence}. 

To address such an issue, orthogonal weight initialization has been extensively studied in the past few years.
For linear networks with arbitrary depth, orthogonality of each composing layers trivially leads to an isometry of the input-output Jacobian, and the benefit over Gaussian initialization in terms of training efficiency has been empirically observed \cite{saxe2013exact} and theoretically justified \cite{hu2020provable}. 
For deep nonlinear networks, isometry may also be achieved if the network works in a local regime that the nonlinearity becomes approximately linear \cite{pennington2018emergence}, and empirical good performance of orthogonal initialization is also observed \cite{mishkin2015all} particularly when combined with proper scaling. 
This line of work culminates at a recent work \cite{xiao2018dynamical} which shows that orthogonal initialization enables effective training of ConvNets with $10,\!000$ layers.
Nonetheless, the performance of the network in \citet{xiao2018dynamical} is far below the state-of-the-art networks, perhaps due to the fact that isometry beyond the initialization point is not guaranteed. 

The Delta initialization adopted in our method is a particular case of orthogonal initialization for convolution kernels. Despite the existence of other orthogonal initialization \cite{xiao2018dynamical}, we advocate the Delta initialization due to its simplicity and good empirical performance in our evaluation for visual recognition tasks.

\subsection{In Context of Weight Regularization}
\label{sec:c_regu}

A plethora of works has explored the idea of regulating the convolution operators in a ConvNet to be orthogonal in the training process \cite{harandi2016generalized,jia2017improving,cisse2017parseval,bansal2018can,zhang2019approximated,li2019efficient,huang2020controllable}. 
It is also found that orthogonal regularization helps with training GANs \cite{brock2018large,liu2019oogan} and RNNs \cite{arjovsky2016unitary,lezcano2019cheap}. 
Despite widely observed performance improvement, the explanation for the effectiveness of orthogonal regularization is rather diverse: it has been justified from the perspective of 
alleviating gradient vanishing or exploding \cite{xie2017all}, 
stabilizing distribution of activation over layers~\cite{huang2018orthogonal}, 
improving generalization~\cite{jia2019orthogonal},
and so on.

In our framework, the benefit of orthogonal regularization lies in that it enforces the network to be close to an isometry. 
To make our argument precise, we first distinguish between the two related concepts of orthogonal weights and orthogonal convolution.
When enforcing orthogonality of convolution kernels in convolutional neural networks, all of the works mentioned above are based on flattening a 4D kernel into a 2D matrix and imposing orthogonality on the matrix.
We refer to such a method as imposing \emph{weight orthogonality}, which is not the same as the orthogonality of convolution as discussed in Section~\ref{sec:convolution}. 
For example, in \citet{bansal2018can} a kernel $\A \in \Re^{M \times C \times k \times k}$ is reshaped into a matrix of shape $M \times (C \times k \times k)$ and is enforced to be row-orthogonal. That is,
\begin{equation}\label{eq:coisometry-conv-flatten}
    \sum_{c=1}^C \langle \text{vec}(\am_{mc}), \text{vec}(\am_{m'c}) \rangle = \begin{cases}
    1 & ~\text{if}~m = m',\\
    0 & ~\text{otherwise},\\
    \end{cases}
\end{equation}
where vec$(\am_{mc})$ denotes a vector of length $k \times k$ obtained from flattening $\am_{mc}$. 
Comparing \eqref{eq:coisometry-conv-flatten} with \eqref{eq:coisometry-conv}, it is clear that such a regularization is necessary, but not sufficient, for the operator $\cA^*$ to be isometric. 
Therefore imposing weight orthogonality as in previous works can be interpreted as partially enforcing network isometry as well.

To the best of our knowledge, the only works that have derived orthogonality for convolution kernels are \citet{xiao2018dynamical} and \citet{wang2019orthogonal}. 
\citet{xiao2018dynamical} provides a means of constructing orthogonal convolution kernels, but it can only represent a subset of all orthogonal convolutions \cite{li2019preventing}. 
In addition, while the method in \citet{xiao2018dynamical} can be used to generate a random orthogonal convolution, it does not provide a means of enforcing orthogonality in the training process. 
We become aware of a very recent work \cite{wang2019orthogonal} during the preparation of this paper, which derives the notion of orthogonal convolution that is equivalent to ours. 
However, the objective of our work is different from theirs. Instead of improving an existing network by adding an additional regularization, our work aims to show that isometry is the most central principle to design effective deep vanilla networks. And as shown in Table~\ref{tab:abl_iso}, using orthogonal regularization alone is not enough for training deep vanilla networks.
Meanwhile, the derivation in \cite{wang2019orthogonal} is based on expressing convolution as matrix-vector multiplication using a doubly block-Toeplitz matrix and using the notion of orthogonality for matrices. 
Such a derivation is limited to 2D convolution while generalization to higher dimensional convolution may become very cumbersome. 
Moreover, the definition is restricted to discrete time signals and does not adapt to continuous time signals. 
In contrast, our definition of orthogonal convolution can be extended for higher-dimensional convolution and for continuous times signals by properly redefining the operator ``*'' in \eqref{eq:def-conv}, therefore may bear broader interests. 

\subsection{In Context of Nonlinear Activation}
\label{sec:c_activation}

Many of the important variants of nonlinear activation functions developed over the past few years can be interpreted as obtaining closer proximity to isometry. 
Early works on improving ReLU such as Leaky ReLU \cite{maas2013rectifier}, Parametric ReLU \cite{he2015delving} and Randomized ReLU \cite{xu2015empirical}, which are generically defined as
\begin{equation}\label{eq:LReLU}
    f(x) =     
    \begin{cases}
      x & \text{if}~~ x \ge 0, \\
      \alpha x  & \text{otherwise},
    \end{cases}
\end{equation}
with the parameter $\alpha >0$ being fixed, learned and randomly chosen, respectively. 
Motivation for this family of activation functions comes from the dead neuron issue of ReLU, which states that negative input values give rise to zero gradients that prevents effective training of the network. 

The Exponential Linear Unit (ELU) \cite{clevert2015fast} and its variants \cite{trottier2017parametric,klambauer2017self}, defined as
\begin{equation}\label{eq:ELU}
    f(x) =     
    \begin{cases}
      \gamma x & \text{if}~~ x \ge 0, \\
      \alpha (\exp(x/\beta)-1)  & \text{otherwise,}
    \end{cases}
\end{equation}
with $\alpha, \beta$ and $\gamma$ being trainable or fixed parameters, is another family of activation functions that may solve the dead neuron issue.
Another important motivation for ELU is that it pushes the mean of the activation closer to zero, therefore alleviates the issue with regular ReLU that the bias of the input signal is always shifted towards positivity. 
Such benefit is further confirmed by a theoretical study \cite{klambauer2017self} which shows that a zero mean and unit variance signal converges towards being zero mean and unit variance after propagating over multiple layers. 
Empirically, a very recent work \cite{huang2019sndcnn} demonstrates that a 50-layer neural network without batch normalization and skip connection may be effectively trained for speech recognition tasks by using ELU instead of ReLU. 

The SReLU that we adopt in the isometric network is closely related to the ELU family. 
To the best of our knowledge, a variant of SReLU (in which the parameter is fixed and not optimizable) first appears in \citet{clevert2015fast}, the paper that proposes ELU, where it is shown to outperform ReLU and Leaky ReLU by a large margin and has similar performance as ELU.
Besides, SReLU is also adopted in \cite{qiu2018frelu} as an important baseline.
Recently, SReLU is rediscovered in \cite{xiang2017effects,singh2019filter} for alleviating the bias shifting issue.

We argue that aside from the benefits claimed by the original papers, the activation function families \eqref{eq:LReLU}, \eqref{eq:ELU} as well as the SReLU are advantageous over the regular ReLU as they bring the network closer to an isometry.
At an intuitive level, both \eqref{eq:LReLU} (with $0 < \alpha < 1$), \eqref{eq:ELU} (with $\gamma=1$ and $0 < \alpha < 1$ and $\beta=1$) and SReLU lie between ReLU and identity, therefore may be considered as a tradeoff between obtaining nonlinearity and isometry. 
From a theoretical perspective, the analysis from \cite{pennington2018emergence} reveals that SReLU combined with random orthogonal initialization can achieve dynamic isometry, while regular ReLU cannot. 

\subsection{In Context of Residual Learning}
\label{sec:c_residual}

Ever since the inception of residual learning \cite{he2016deep}, there has been no lack of effort in further enhancing its performance. In particular, there is a line of work that demonstrates the benefit of reducing the energy on the residual branch of a ResNet. This is studied from different perspectives such as dynamic isometry~\cite{tarnowski2019dynamical,qiu2018frelu}, training failure modes~\cite{taki2017deep,hanin2018start,balduzzi2017shattered}. Such design is also empirically applied in different ways such as BatchNorm initialization~\cite{goyal2017accurate}, adding a small scalar in the residual branch~\cite{zagoruyko2017diracnets,szegedy2017inception}, and 0-initialized convolution~\cite{zhang2019fixup}, all of which brings the network closer to an isometry. With a careful design along this line of study \cite{zhang2019fixup}, deep ResNet can be effectively trained to obtain competitive performance on visual recognition tasks, even without the help of BatchNorm layers. This line of research as well as our simple R-ISONet verifies the effectiveness of isometry in the design the network structures of residual learning.

\section{Conclusion}
In this paper, we have demonstrated through a principle-guided design and strong empirical evidence why isometry is likely to be the main key property that enables effective learning of deep networks and ensures high performance on real visual recognition tasks. With this design principle, one may achieve competitive performance with much simplified networks and eased training. We also argue that isometric learning provides a unified principle that helps explain numerous  ideas, heuristics and regularizations scattered in the literature that exploit this property and are found effective. We believe that the isometry principle may help people design or discover new simple network operators and architectures with much-improved performance in the future. 

\section*{Acknowledgements}
The authors acknowledge support from Tsinghua-Berkeley Shenzhen Institute Research Fund.
Haozhi is supported in part by DARPA Machine Common Sense. Xiaolong is supported in part by DARPA Learning with Less Labels. 
We thank Yaodong Yu and Yichao Zhou for insightful discussions on orthogonal convolution. We would also like to thank the members of BAIR for fruitful discussions and comments.

\clearpage

\bibliography{main}
\bibliographystyle{icml2020}

\clearpage

\begin{appendices}

\section{Proof to Theorem~\ref{thm:convolution_isometry}}\label{ap:proof}

\begin{proof}
From Definition~\ref{def:isometry}, $\cA$ is an isometry if and only if
\begin{equation}\label{eq:prf-isometry-def}
    \langle \cA \x, \cA \x' \rangle = \langle \x, \x' \rangle, ~~\forall \{\x, \x'\} \subseteq \Re^{C\times H\times W}.
\end{equation}
By the property of the adjoint operator, \eqref{eq:prf-isometry-def} is equivalent to$\!\!\!$
\begin{equation}
     \langle \cA^* \cA \, \x, \,\x' \rangle = \langle \x, \x' \rangle, ~~\forall \{\x, \x'\} \subseteq \Re^{C\times H\times W},
\end{equation}
which holds if and only if 
\begin{equation}
    \cA^* \cA\, \x = \x, ~~\forall\x = (\xm_1, \ldots, \xm_C) \in \Re^{C\times H\times W}.
\end{equation}
By using \eqref{eq:conv_A} and \eqref{eq:conv_At}, we rewrite $\cA^* \cA \,\x$ as
\begin{multline}\label{eq:prf-AtA}
    \sum_{m=1}^M \sum_{c=1}^C\Big(\am_{m1} * \big( \am_{mc} \star \xm_c\big), \ldots, \am_{mC} * \big(\am_{mc} \star \xm_c \big)\Big)\\
    = \sum_{m=1}^M \sum_{c=1}^C\Big(\big(\am_{mc} \star \am_{m1}\big) * \xm_c, \ldots, \big(\am_{mc} \star \am_{mC}\big) * \xm_c \Big).
\end{multline}

In \eqref{eq:prf-AtA}, we have used the fact that for arbitrary 2D signals $\am, \am'$ and $\xm$, 
\begin{multline}\label{eq:prf-conv-corr}
    \am * (\am' \star \xm) = (\am' \star \xm) * \am \\= \am' \star (\am * \xm) = \am' \star (\am * \xm) = (\am' \star \am) * \xm,
\end{multline}
which follows from the commutative property of convolution (i.e., $\am * \xm = \xm * \am$) and the associative property of convolution and correlation (i.e., $\am \star (\am' * \xm) = (\am \star \am') * \xm$). 
By equating the last line of \eqref{eq:prf-AtA} with $\x = (\xm_1, \ldots, \xm_C)$, we get that $\cA$ is an isometry if and only if
\begin{equation}
    \sum_{m=1}^M \sum_{c=1}^C \big(\am_{mc} \star \am_{mc'}\big) * \xm_c = \xm_{c'} ~~\forall \c' \in \{1, \ldots, C\}
\end{equation}
holds for all $\x = (\xm_1, \ldots, \xm_C)$, which is equivalent to \eqref{eq:isometry-conv}.
Analogously, we can show that $\cA^*$ is an isometry if and only if \eqref{eq:coisometry-conv} holds.
\end{proof}

\section{Additional Experiments}\label{ap:experiments}

\textbf{Isometric Components on ResNet.} The isometric components are not heuristic engineering components but are specifically designed to promote neural network's isometric property. To validate this hypothesis, we show that naively adding those components to a standard ResNet34 will only have marginal or even negative impact on its performance (see Table ~\ref{tab:abl_resnet}). Standard ResNet, which has the BatchNorm layers, is not so sensitive to Delta initialization, as shown in Table ~\ref{tab:abl_resnet} (b). Applying SReLU will however even decrease the performance since the nonlinearity effect is compromised by BatchNorm due to its effect of forcing output to have zero mean. In addition, imposing orthogonal regularization improves the performance by only a small margin.

\begin{table}
    \centering
    \small
    \setlength{\tabcolsep}{3pt}
    \renewcommand{\arraystretch}{1.4}
    \begin{tabular}{c|c|ccc|c}
    & Method & SReLU & {\begin{mytabular}[1.2]{c} Delta \\ Init. \end{mytabular}} & \begin{mytabular}[1.2]{c} Ortho. \\ Reg. \end{mytabular} & \begin{mytabular}[1.2]{c} Top-1 \\ Accuracy (\%) \end{mytabular}  \\
    \hline
    \hline
    (a) & \multirow{5}{*}{ResNet} & & & & 73.29 \\
    (b) & & \checkmark & & & 72.92 \\
    (c) & & & \checkmark & & 71.80 \\
    (d) & & & & \checkmark & 73.60 \\
    (e) & & \checkmark & \checkmark & \checkmark & 71.53
    \end{tabular}
    \caption{ImageNet Top-1 accuracy for standard ResNet34 with additional isometric components. 
    The results show that these components are specifically designed for imposing isometry of the network, instead of heuristic engineering components.}
    \label{tab:abl_resnet}
\end{table}

\end{appendices}








\end{document}